\documentclass[10pt,twocolumn,letterpaper]{article}

\usepackage[pagenumbers]{cvpr} 

\usepackage[accsupp]{axessibility} 

\usepackage{graphicx}
\usepackage{amsmath}
\usepackage{amssymb}
\usepackage{booktabs}

\usepackage{array,multirow,graphicx}
\usepackage{float}
\usepackage{amsmath}
\usepackage{mathtools}
\usepackage{bbm}
\usepackage{hhline}
\usepackage{boldline}

\usepackage[pagebackref,breaklinks,colorlinks]{hyperref}

\usepackage[capitalize]{cleveref}
\crefname{section}{Sec.}{Secs.}
\Crefname{section}{Section}{Sections}
\Crefname{table}{Table}{Tables}
\crefname{table}{Tab.}{Tabs.}

\def\cvprPaperID{6933} 
\def\confName{CVPR}
\def\confYear{2023}

\begin{document}



\title{Leveraging Hidden Positives for Unsupervised Semantic Segmentation}

\author{Hyun Seok Seong, WonJun Moon, SuBeen Lee, Jae-Pil Heo\thanks{Corresponding author} \\
Sungkyunkwan University\\
{\tt\small \{gustjrdl95, wjun0830, leesb7426, jaepilheo\}@skku.edu}
}


\maketitle

\begin{abstract}
Dramatic demand for manpower to label pixel-level annotations triggered the advent of unsupervised semantic segmentation.
Although the recent work employing the vision transformer~(ViT) backbone shows exceptional performance, there is still a lack of consideration for task-specific training guidance and local semantic consistency.
To tackle these issues, we leverage contrastive learning by excavating hidden positives to learn rich semantic relationships and ensure semantic consistency in local regions.
Specifically, we first discover two types of global hidden positives, task-agnostic and task-specific ones for each anchor based on the feature similarities defined by a fixed pre-trained backbone and a segmentation head-in-training, respectively.
A gradual increase in the contribution of the latter induces the model to capture task-specific semantic features.
In addition, we introduce a gradient propagation strategy to learn semantic consistency between adjacent patches, under the inherent premise that nearby patches are highly likely to possess the same semantics.
Specifically,
we add the loss propagating to local hidden positives, semantically similar nearby patches, in proportion to the predefined similarity scores.
With these training schemes, our proposed method achieves new state-of-the-art~(SOTA) results in COCO-stuff, Cityscapes, and Potsdam-3 datasets.
Our code is available at: \href{https://github.com/hynnsk/HP}{https://github.com/hynnsk/HP}.
\end{abstract}

\section{Introduction}
\label{sec_introduction}


\begin{figure}[t]
    \centering
    \includegraphics[width=0.47\textwidth]{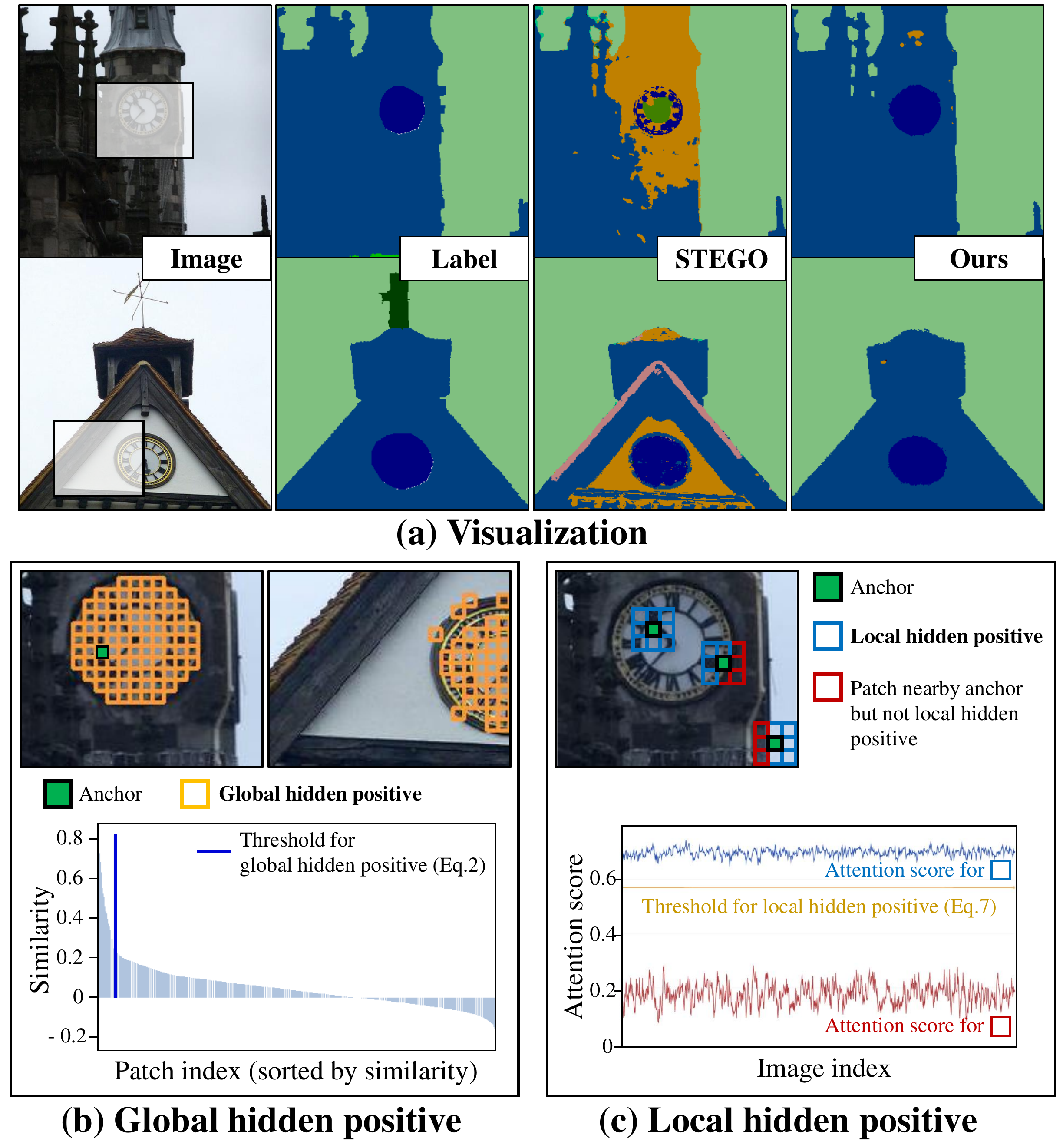}
    \caption{
    Assuming a mini-batch comprising two images shown in (a), we describe two types of hidden positives, to leverage for contrastive learning.
    \textbf{(a)}~With two types of hidden positives introduced in (b) and (c), we provide an example of how our training scheme provides more precise and consistent semantics.
    \textbf{(b)}~(top)~Semantically analogous patches throughout the mini-batch are selected as global hidden positives. 
    (bottom)~Data-driven criterion per anchor is designed for reliable positive collection.
    With the criterion, selected positives are illustrated in (b~top).
    \textbf{(c)}~(top)~We define local hidden positives for each anchor to be the adjacent patches with high semantic consistency, i.e., blue boxes.
    (bottom)~Average attention scores for adjacent patches from the pretrained transformer architecture. 
    The blue line represents the attention score for local hidden positives while the red line for patches neighboring anchors but having low semantic consistency.
    }
    \label{fig_motivation}
\end{figure}

Semantic segmentation is a major task for scene understanding and plays a crucial role in many applications including medical imaging and autonomous driving~\cite{unet, deeplab, danet, setr, segformer, ocr}.
However, existing supervised approaches demand large-scale pixel-level annotations which require huge labeling costs.
It has triggered the advent of weakly-supervised~\cite{ficklenet, seam, nsrom, mctformer} and unsupervised semantic segmentation~\cite{iic, picie, transfgu, stego} which are to learn without expensive pixel-level annotations.

Particularly, unsupervised semantic segmentation is one of the most challenging tasks, since it needs to capture pixel-level semantics from unlabeled data.
In this context, clustering-based approaches have been proposed to learn semantic-preserving clusters by attracting the augmented views in the pixel-level~\cite{iic, picie}.
They implemented the intuition of contrastive learning~\cite{simclr, moco, swav, byol, pirl, simsiam} by ensuring the augmented pairs yield symmetric cluster assignments.
More recently, as discovering pixel-level semantics from scratch is challenging, STEGO~\cite{stego} broke down the problem into learning the representation and learning the segmentation head.
With the learned patch-level representation from the seminal work in unsupervised learning~\cite{dino, selfpatch}, they train the segmentation head with a distillation strategy.
Although they have made great advancements, we point out their limitations in that they rely solely on a fixed backbone that is not specifically trained for the segmentation task and overlook the importance of semantic consistency along the adjacency that could be a crucial clue for segmentation.
To take these into consideration, we leverage contrastive learning based on the mined hidden positives to ensure contextual consistency along the patches with analogous semantics, particularly the nearby patches, as described in Fig.~\ref{fig_motivation}.
Specifically, we elaborately select the pseudo-positive samples~(i.e., global hidden positive, GHP) for contrastive learning to learn semantic consistency.
Also, to ensure local consistency, we propagate the loss gradient to the adjacent patches~(i.e., local hidden positive, LHP) in proportion to their equivalency.
First, the GHP selection process is designed with two types of data reference pools, task-agnostic and task-specific, to collect the semantically consistent patch features throughout the mini-batch per anchor.
For instance, the task-agnostic data reference pool is composed of features extracted by the unsupervised pretrained backbone.
On the other hand, the task-specific reference pool is constructed with the features from the segmentation head-in-training to complement task relevance.
Based on the two reference pools, two sets of GHP are selected each with generalized and task-specific perspectives.
Second, to implement the property of locality and prevent the semantics from fluctuating, we propagate the loss gradient to adjacent patches~(i.e., LHP) in proportion to the similarity scores built within the pretrained backbone.
This enables the model to learn the relevance of the local context that nearby patches often belong to the same instance.

Our main contributions are summarized as:
\begin{itemize}
\item{We propose a novel method to discover semantically similar pairs, called global hidden positives, to explicitly learn the semantic relationship among patches for unsupervised semantic segmentation.}
\item{We utilize the task-specific features from a model-in-training and validate the effectiveness of progressive increase of their contribution.}
\item{A gradient propagation to nearby similar patches, local hidden positives, is developed to learn local semantic consistency which is the nature of segmentation.}
\item{Our approach outperforms existing state-of-the-art methods across extensive experiments.}
\end{itemize}
\section{Related Work}

\subsection{Unsupervised Semantic Segmentation}
Semantic segmentation has been extensively studied for its wide applicability~\cite{deeplab, danet, setr, segformer, ocr, segmenter}, but collecting pixel-level annotations requires expensive costs.
Therefore, many studies~\cite{iic, picie, transfgu, stego, segsort, maskcontrast} attempted to address semantic segmentation without any supervision.
Earlier techniques tried to learn semantic correspondence at the pixel level.
IIC~\cite{iic} maximizes the mutual information between the features of two differently augmented images, and PiCIE~\cite{picie} learns photometric and geometric invariances as an inductive bias.
Yet, their training process highly depends on data augmentation, and learning semantic consistency without any prior knowledge is challenging.
Therefore, recent methods~\cite{transfgu, stego} adopted the ViT model trained in a self-supervised manner, i.e., DINO~\cite{dino}, as a backbone architecture.
For instance, TransFGU~\cite{transfgu} relocates the high-level semantic features from DINO into low-level pixel-wise features by generating pixel-wise pseudo labels.
On the other hand, STEGO~\cite{stego} utilizes knowledge distillation that learns correspondences between features extracted from DINO.
Although STEGO shows a dramatic performance improvement compared to the prior works, it heavily relies on the pretrained backbone and overlooks the property of local consistency that the adjacent pixels are likely to belong to the same category.
On the other hand, our training is driven by both the task-agnostic and task-specific pseudo-positive features, and the gradients are conditionally propagated to the neighboring patches, thereby ensuring task-specificity and locality.

\begin{figure*}[t]
    \centering
    \includegraphics[width=0.98\textwidth]{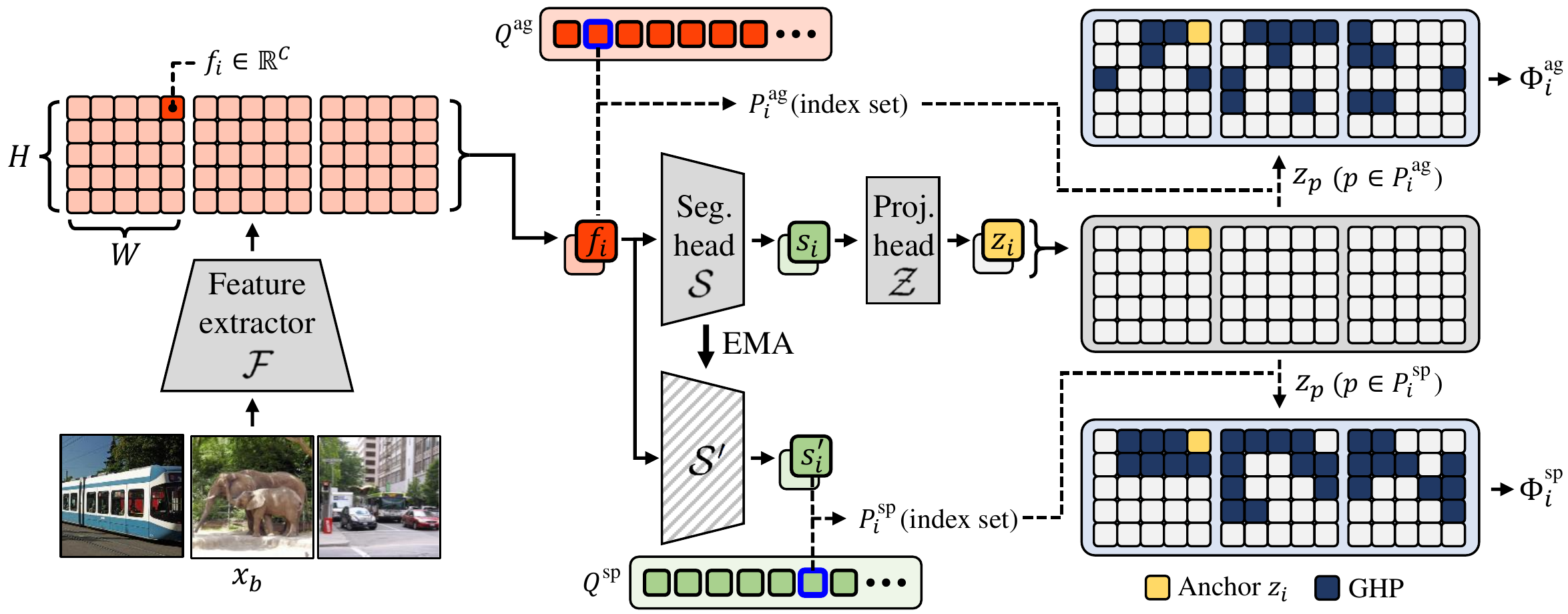}
    \caption{
    Illustration of the global hidden positive~(GHP) selection process. 
    Our GHP can be divided into two sub-sets: task-agnostic and task-specific.
    An index set of task-agnostic GHP ${P}_{i}^{\text{ag}}$ comprises the indices of positives discovered within the task-agnostic reference pool $Q^{\text{ag}}$.
    Note that, $Q^{\text{ag}}$ is composed of randomly sampled features extracted by the feature extractor $\mathcal{F}$.
    Once the anchor feature $f_i$ is projected to $z_i$, other patches in the mini-batch are gathered as positives if their similarity with the anchor feature exceeds the similarity between the anchor and the most similar feature in $Q^{\text{ag}}$.
    On the other hand, task-specific GHP is discovered in a similar manner but with task-specific reference pool $Q^{\text{sp}}$ which keeps being updated with the features from the momentum segmentation head $S'$.
    Whereas the task-agnostic GHP set solely contributes to the initial training, the task-specific GHP set gradually replaces the portion of the task-agnostic set until the end of training.
    }
    \label{fig_main}
\end{figure*}

\subsection{Contrastive Learning}
Self-supervised learning methods~\cite{rotnet, deepcluster, simclr, moco, swav, byol, simsiam} aim to learn general representations without any annotations. 
Thanks to their good representation capability, they have been employed to yield remarkable performances in various downstream tasks~\cite{picie, hendrycks2019using, supcon, lorot, supseg}.
Among them, contrastive learning approaches~\cite{simclr, moco, swav, byol, pirl, simsiam} have shown unrivaled performances.
In general, they learn representation by attracting a self-augmented set and repulsing other images~\cite{simclr, swav}.
Other variants use additional memory~\cite{moco} or only exploit the positive set for the attraction~\cite{byol, simsiam}.
This training scheme has also been utilized for unsupervised semantic segmentation~\cite{iic, picie}.
However, the aforementioned methods only considered augmented pairs for the positive which makes them very sensitive to the quality of the augmentation techniques.
Our work differs in that our key idea is to collect and exploit reliable pseudo-positives throughout the mini-batch as described in Fig.~\ref{fig_motivation}~(b).


\section{Method}






In this section, we introduce the pseudo-positive selection strategy to discover hidden positives with analogous semantics in Sec.~\ref{sec_global_positive}, the training objective with discovered positives in Sec.~\ref{sec_objective_function}, and the gradient propagation scheme to preserve the property of locality in Sec.~\ref{sec_loss_propagation}.



\subsection{Preliminary}
In unsupervised semantic segmentation, the model utilizes unlabeled image set $X=\{x_b\}_{b=1}^B$ where $B$ is the number of training data in the mini-batch. 
Given an image $x_b$ processed to the feature extractor $\mathcal{F}$, we have $H \cdot W$ features of $f_{i}\in \mathbb{R}^{C}$, where $i \in [ 1 , ... , H\cdot W ]$.
Subsequently, the segmentation head $\mathcal{S}$ maps a patch feature $f_i$ to the corresponding segmentation feature $s_{i}\in \mathbb{R}^{K}$.
And then, the projection head $\mathcal{Z}$ produces a projected vector $z_{i}\in\mathbb{R}^{K}$ to formulate a contrastive loss function.
In the inference stage, we use the segmentation feature $s_i$.


Based on the projected vector $z_i$ for the $i$-th patch, let $j$ be the index of the augmented patch of $i$-th one.
Then, the conventional self-supervised contrastive loss~\cite{simclr} for $i$-th patch in unsupervised semantic segmentation can be defined as follows:
\begin{equation}
\label{eq_contrastive_loss_self}
    L^{\text{self}}_i = - \text{log} \frac{\text{exp}(\text{sim}(z_i,z_j / \tau))}{\sum_{a \in A}  \text{exp}(\text{sim}(z_i,z_a) / \tau)},
\end{equation}
where $A$ indicates a set of all indexes except $i$, $\tau$ denotes the scalar temperature parameter, and $\text{sim}(\cdot,\cdot)$ is cosine similarity between two vectors.


\subsection{Global Hidden Positives}
\label{sec_global_positive}
Learning mutual information with augmented pixels only provides insufficient training signal in unsupervised semantic segmentation~\cite{stego}.
Therefore, it is important to discover hidden pseudo-positives to tailor the contrastive loss for unsupervised segmentation.
To discover the hidden positives at the initial stage, we utilize the self-supervised pretrained backbone~\cite{dino} as the task-agnostic criterion.
Then, we gradually increase the contribution of hidden positives found in a task-specific way for the training.
Fig.~\ref{fig_main} provides the overview of the global hidden positive~(GHP) selection process.

Initially, the pretrained backbone is utilized to construct a task-agnostic reference pool to assess whether other features in the mini-batch are semantically-alike for each anchor feature.
Specifically, task-agnostic reference pool, $Q^{\text{ag}} = \{q_{m}\}_{m=1}^{M}$, is composed of $M$ randomly sampled features that are extracted by the unsupervised pretrained backbone $\mathcal{F}$. 
Note that, we only sample a single patch feature per image to ensure the semantic randomness of the reference pool.
This task-agnostic reference pool is fixed as the pretrained backbone is frozen throughout the training. 

Once the reference pool is gathered, for each patch feature $f_i$, we define an anchor-dependent similarity criterion $c_i$ to collect positives, as the distance to the closest feature within the reference pool $Q^{\text{ag}}$ by the cosine similarity:
\begin{equation}
\label{eq_select_criterion}
    c_{i} = \max_{q_m\in Q^\text{ag}} \text{sim}(q_m, f_{i}).
\end{equation}
For each anchor feature $f_i$, we basically treat the other feature in the mini-batch $f_j$ as positive if the similarity between $f_i$ and $f_j$ is greater than $c_i$.
Still, although one patch feature might be the positive sample for the other, it may not hold mutually.
This is because the criterion $c_i$ is anchor-dependent.
To endow consistency in training, we make the GHP selection symmetric to prevent the relation between two patches from being ambiguous.
Therefore, index set of GHP ${P}^{\text{ag}}_{i}$ for each $i$-th anchor feature $f_{i}$ is defined as follows:
\begin{equation}
\label{eq_positive_pre_index_rewritten}
    {P}_{i}^{\text{ag}} = \{j \mid \text{sim}(f_{i}, f_{j}) > c_{i}
    \;\vee\; \text{sim}(f_{i}, f_{j}) > c_{j}\},
\end{equation}
where $j$ indicates the index for different patch features in the mini-batch.
Accordingly, such a distribution-aware reference pool allows the discovery of globally analogous features in consideration of each anchor.

However, although the reference pool built upon the features from an unsupervised pretrained network can serve as an appropriate basis for positivity, it may be insufficient since it lacks task-specificity. 
We argue that features from the segmentation head are more task-specific than those from the pretrained backbone.
Therefore, along with the GHP selected by $P^{\text{ag}}$, we construct additional task-specific GHP utilizing the features from the segmentation head.

Specifically, an index set of task-specific GHP $P_{i}^{\text{sp}}$ is formed similarly to Eq.~\ref{eq_positive_pre_index_rewritten} by comparing the features $s^\prime=\mathcal{S}^\prime(f)$ and task-specific reference pool $Q^{\text{sp}}$,
where ${\mathcal{S}^\prime}$ indicates the momentum segmentation head, and the task-specific reference pool $Q^{\text{sp}}$ comprises $s^\prime$.
Also, along with the update of the segmentation head, during training, the reference pool is periodically renewed.
Formally, with $c^\prime_{i}$, calculated by substituting $Q^\text{ag}$ and $f_i$ in Eq.~\ref{eq_select_criterion} with $Q^\text{sp}$ and $s^\prime_i$, respectively, the $P_{i}^{\text{sp}}$ is expressed as follows:
\begin{equation}
    P_{i}^{\text{sp}} = \{j \mid \text{sim}({s^\prime}_i, {s^\prime}_j) > c^\prime_{i} \vee \text{sim}({s^\prime}_i, {s^\prime}_j) > c^\prime_{j} \}.
\end{equation}
Note that, the usage of the momentum segmentation head is for the stability of the reference pool~\cite{ema, sessd}.


\begin{figure*}[t]
    \centering
    \includegraphics[width=1.\textwidth]{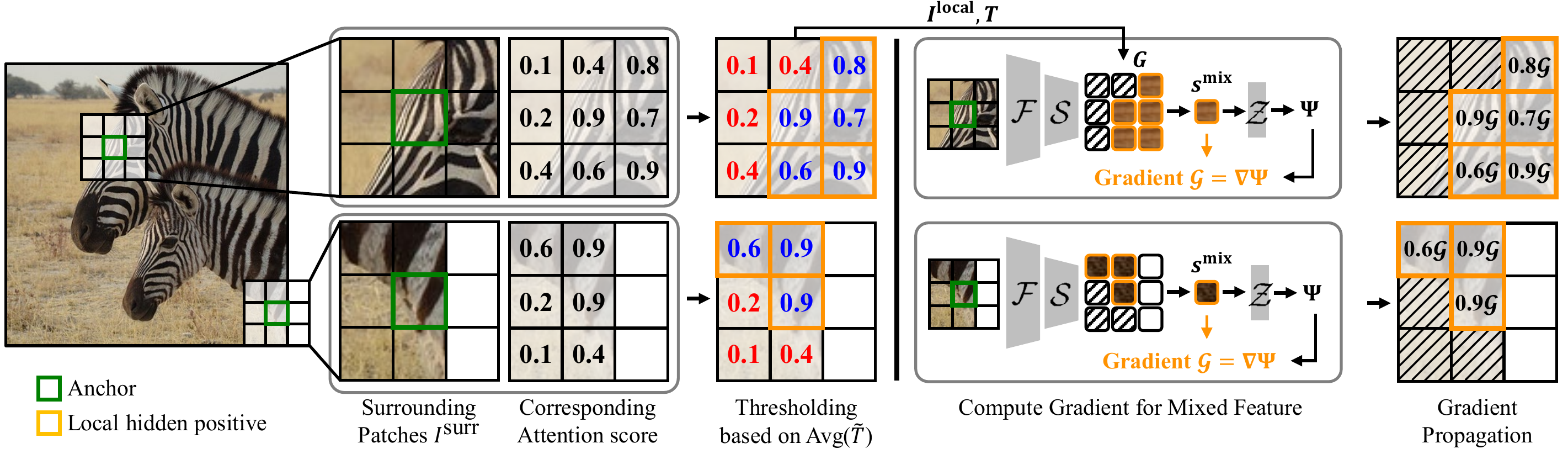}
    \caption{
    Illustration of our gradient propagation strategy to preserve local semantic consistency. 
    For each anchor, with its surrounding patches $I^{\text{surr}}$ and corresponding attention scores from feature extractor $\mathcal{F}$, local hidden positives~(LHP) $I^{\text{local}}$ are appointed based on the threshold Avg($\tilde{T}$)~(Eq.~\ref{eq_local_positive_index}).
    In a forward pass, the features of LHP $G$~(Eq.~\ref{eq_g_and_t}) are mixed by weighted average according to the attention scores $T$ to compute the objective function $\Psi$. 
    In this way, the loss gradient propagates toward the LHP in proportion to $T$ in the backward pass.
    }
    \label{fig_gradient_propagation}
\end{figure*}

\subsection{Objective Function}
\label{sec_objective_function}
To formulate a contrastive objective with mined GHP in Sec.~\ref{sec_global_positive}, we also need negative features.
As we collected the positives throughout the mini-batch, the naive implementation of contrastive learning would utilize all features except the selected positives in the mini-batch as the negatives.
However, since an immoderate increase in the size of the negative set may disturb the model training~\cite{eisnet}, we form a negative set $N_i$ by randomly choosing $\rho\%$ of the remaining patches for each $i$-th anchor. Note that, the separate index sets of negative samples $N_i^{\text{ag}}$ and $N_i^{\text{sp}}$ are defined for each $P_i^{\text{ag}}$ and $P_i^{\text{sp}}$, correspondingly.
Also, unlike Eq.~\ref{eq_contrastive_loss_self}, our contrastive loss for each $i$-th anchor is more like a supervised objective~\cite{supcon} since we are given multiple positives:
\begin{eqnarray}
\label{eq_contrastive_loss}
    L^{\text{cont}}(z_i, P, N) = \frac{-1}{|P|}\sum_{p \in P} \text{log} \frac{\text{exp}(\text{sim}(z_i,z_p) / \tau)}{\sum\limits_{n \in (N \cup P)} \text{exp}(\text{sim}(z_i,z_n) / \tau)},
\end{eqnarray}


where $z_i$, $P$, and $N$ are the projected anchor vector $\mathcal{Z}(\mathcal{S}(f_i))$, positive index set, and negative index set, respectively.
For simplicity, we use $\Phi_{i}^{\text{ag}}$ and $\Phi_{i}^{\text{sp}}$ to denote the objective functions with task-agnostic GHP $P^{\text{ag}}_i$ and task-specific GHP $P^{\text{sp}}_i$ for each $i$-th anchor as follows:
\begin{equation}
\label{eq_contrastive_loss_global_task_agnostic}
\begin{split}
    \Phi_{i}^{\text{ag}} &= L^{\text{cont}}(z_i, P^{\text{ag}}_i, N^{\text{ag}}_{i})
    \\
    \Phi_{i}^{\text{sp}} &= L^{\text{cont}}(z_i, P^{\text{sp}}_i, N^{\text{sp}}_{i}).
\end{split}
\end{equation}

\subsection{Gradient Propagation to Local Hidden Positives}
\label{sec_loss_propagation}
Besides considering the semantically analogous features globally, it is a common hypothesis that nearby pixels are highly likely to belong to the same semantic class.
To this end, we consider the property of locality by propagating the loss gradient to the surrounding features of the anchor.
Still, the propagation should be cautiously designed since semantic labels of the adjacent patches are not given; semantic consistency between adjacent patches mostly holds, but sometimes does not (i.e., at object boundaries).
Thus, to decide the semantically consistent patches nearby, we utilize the attention scores from the unsupervised pretrained ViT backbone $\mathcal{F}$.




In detail, we first define the index set $I_i^\text{surr}$ for surrounding patches of the $i$-th anchor including $i$-th anchor itself. 
Also, given the spatial attention score for $i$-th anchor ${\tilde{T}}_{i} \in \mathbb{R}^{H\cdot W}$ from the last self-attention layer in the backbone $\mathcal{F}$, we use the average value of $\tilde{T}_{i}$ as a threshold to select an index set of LHP $I^{\text{local}}_i$ among $I^{\text{surr}}_i$:
\begin{equation}
    \label{eq_local_positive_index}
    I^{\text{local}}_i = \{j \mid j \in I^{\text{surr}}_i \;\land\; t_{j} > \text{Avg}(\tilde{T}_{i})\},
\end{equation}
where $t_{j}$ is $j$-th element of $\tilde{T}_i$ and Avg($\cdot$) denotes the averaging function. 
The visualization of the attention scores for the adjacent patches is shown in Fig.~\ref{fig_motivation}~(b). 
By utilizing $I^{\text{local}}_i$, we obtain the surrounding positive features $G_i$ and the corresponding attention score $T_{i}$ for LHP as follows:
\begin{equation}
\label{eq_g_and_t}
\begin{split}
    G_{i} &=\{s_{j} \mid j \in I^{\text{local}}_i \}
    \\
    T_i &= \{t_j \mid j \in I^{\text{local}}_i\},
\end{split}
\end{equation}
where $s_j$ is the feature of $j$-th patch extracted by the segmentation head $\mathcal{S}$.

To propagate the loss gradient to the LHP, we mix the patch features in $G_i$ in proportion to the corresponding attention scores in $T_i$.
Formally, the mixed patch composed of LHP $s^{\text{mix}}_{i}$ is expressed as:
\begin{equation}
    s^{\text{mix}}_{i} = \frac{1}{|I_{i}^{\text{local}}|} \sum_{j \in I_{i}^{\text{local}}} \sigma g_{j}t_{j},
\end{equation}
where $\sigma$ is a scalar value to scale the attention score, and $g_j$ and $t_j$ indicate the $j$-th element of $G_i$ and $T_i$, respectively.
Then, we define the objective functions $\Psi_{i}^{\text{ag}}$ and $\Psi_{i}^{\text{sp}}$ to learn the locality by inserting the projected mixed vector $z^{\text{mix}}_i = \mathcal{Z}(s_i^{\text{mix}})$ into Eq.~\ref{eq_contrastive_loss} as follows:
\begin{equation}
\label{eq_contrastive_loss_local_tas_agnostic}
\begin{split}
    \Psi_{i}^{\text{ag}} &= L^{\text{cont}}( z^{\text{mix}}_i, P^{\text{ag}}_i, N^{\text{ag}}_i )
    \\
    \Psi_{i}^{\text{sp}} &= L^{\text{cont}}( z^{\text{mix}}_i, P^{\text{sp}}_i, N^{\text{sp}}_i ).
\end{split}
\end{equation}
Since these functions are calculated by utilizing the mixed vector $z^\text{mix}_i$, the loss gradients are propagated to all features composing the $G_i$, as described in Fig.~\ref{fig_gradient_propagation}.
Therefore, the semantically-alike surrounding vectors in $G_i$ are updated in the same direction, thereby retaining semantic consistency within the neighboring patches.

Overall, by combining all loss formulations with consistency regularizer $R_i$ that minimizes Euclidean distance between the projected vectors of two differently augmented patches, the final loss function is defined as follows:
\begin{equation}
\label{eq_contrastive_loss_local_overall}
    L_i = (\Phi_i^{\text{ag}} + \Psi_i^{\text{ag}}) + 
    \lambda(\Phi_i^{\text{sp}}+\Psi_i^{\text{sp}})+\alpha R_i,
\end{equation}
where $\lambda$ and $\alpha$ control the contribution of each loss. For instance, $\lambda$ gradually increases from $0$ to $1$ during the training and $\alpha$ remains constant at $0.05$ throughout the training.
Note that, the sample with zero positive is excluded from training although it rarely exists.



\section{Experiments}

\subsection{Datasets and Experimental Settings}
\paragraph{Datasets.}
We utilize COCO-stuff~\cite{coco}, Cityscapes~\cite{cityscapes}, and Potsdam-3 datasets following the existing works~\cite{stego, picie, iic}.
COCO-stuff is a large-scale scene understanding dataset that consists of dense pixel-level annotations and Cityscapes is a more recently publicized dataset having street scenes across 50 different cities.
Potsdam-3 dataset contains satellite images.
Following the baselines~\cite{iic, picie, stego}, we choose the 27 classes for COCO-stuff and Cityscapes datasets, and 3 classes for Potsdam-3 dataset.


\paragraph{Evaluation Protocols and Metrics.}
To evaluate our approach, we conduct two testing methods; clustering and linear probe~\cite{stego}.
Clustering is to measure how well the semantic-preserving clusters are formed.
Once the unlabeled clusters are computed with the extracted representations, clusters are matched with the ground truth class labels using the Hungarian matching algorithm.
On the other hand, the linear probe is a popular method to evaluate the quality of representations learned in unsupervised manners.
Specifically, an additional linear layer is learned with representations extracted from the frozen unsupervised model for evaluation and ground truth labels.
Inferring the representations of test data extracted by the frozen model with the learned linear layer, we can measure the quality of representations.
With two types of protocols, the performance is measured by two common metrics;  accuracy~(Acc.) and mean Intersection Over Union~(mIoU).

\begingroup
\setlength{\tabcolsep}{4.2pt} 
\renewcommand{\arraystretch}{1.0} 
\begin{table}[t!]
    \centering
    \small
    \begin{tabular}{l c c c  c c}
        \hlineB{2.5}
        \multirow{2}{*}{Method}& \multirow{2}{*}{Backbone} & \multicolumn{2}{c}{Unsupervised} & \multicolumn{2}{c}{Linear} \\
        \multicolumn{2}{l}{}  & Acc. & mIoU & Acc. & mIoU \\
        \hlineB{2.5}
        DC~\cite{deepcluster} & R18+FPN & 19.9 & - & - & - \\
        MDC~\cite{deepcluster} & R18+FPN & 32.2 & 9.8 & 48.6 & 13.3 \\
        IIC~\cite{iic} & R18+FPN & 21.8 & 6.7 & 44.5 & 8.4 \\
        PiCIE~\cite{picie} & R18+FPN & 48.1 & 13.8 & 54.2 & 13.9 \\
        PiCIE+H~\cite{picie} & R18+FPN & 50.0 & 14.4 & 54.8 & 14.8 \\
        \hline
        DINO~\cite{dino} & ViT-S/8 & 28.7 & 11.3 & 68.6 & 33.9 \\ 
        + TransFGU~\cite{transfgu} & ViT-S/8 & 52.7 & 17.5 & - & - \\
        + STEGO~\cite{stego} & ViT-S/8 & 48.3 & 24.5 & 74.4 & 38.3 \\
        + HP~(Ours) & ViT-S/8 & \textbf{57.2} & \textbf{24.6} & \textbf{75.6} & \textbf{42.7} \\ 
        \hline
        DINO~\cite{dino} & ViT-S/16 & 22.0 & 8.0 & 50.3 & 18.1 \\ 
        + STEGO~\cite{stego} & ViT-S/16 & 52.5 & 23.7 & 70.6 & 34.5 \\ 
        + HP~(Ours) & ViT-S/16 & \textbf{54.5} & \textbf{24.3} & \textbf{74.1} & \textbf{39.1} \\ 
        \hline
        SelfPatch~\cite{selfpatch} & ViT-S/16 & 35.1 & 12.3 & 64.4 & 28.5 \\ 
        + STEGO~\cite{stego} & ViT-S/16 & 52.4 & 22.2 & 72.2 & 36.0 \\ 
        + HP~(Ours) & ViT-S/16 & \textbf{56.1} & \textbf{23.2} & \textbf{74.9} & \textbf{41.3} \\ 
        \hlineB{2.5}
    \end{tabular}
    \caption{Experimental results on COCO-stuff dataset with various backbones and pretrained models.}
    \label{Tab.coco}
\end{table}
\endgroup

\paragraph{Implementation Details.}
For fair comparisons with our baselines~\cite{stego, transfgu},
we follow them to mainly use DINO pretrained ViT models as a backbone network $\mathcal{F}$ for the COCO-stuff dataset.
In addition, we also test with the advanced backbone, SelfPatch~\cite{selfpatch}, on the COCO-stuff dataset.
The segmentation head $\mathcal{S}$ is constructed with a two-layer RELU MLP as STEGO, and the projection head $\mathcal{Z}$ is composed of a linear layer equipped with a normalization layer.
The embedding dimension $K$ is set to 512 for ViT-S/8 and ViT-B/8 models, and 256 for ViT-S/16.
We train the model for 3, 20, and 10 epochs for COCO-stuff, Cityscapes, and Potsdam-3 datasets, respectively, based on the AdamW optimizer with a learning rate of 0.0005 and weight decay of 0.1.
The task-specific reference pool $Q^{\text{sp}}$ is renewed every 100 iterations throughout the training.
The percentage of negative samples usage $\rho$ is set to 2.
In the last stage, we add a feature refinement step utilizing Conditional Random Field~\cite{crf} as did in STEGO.
Evaluation metrics, i.e., clustering and linear probe, are optimized with the Adam optimizer each with learning rates of 0.005 and 0.001.
\begingroup
\setlength{\tabcolsep}{4.2pt} 
\renewcommand{\arraystretch}{1.0} 
\begin{table}[!t]
    \centering
    \small
    \begin{tabular}{l c c c  c c}
        \hlineB{2.5}
        \multirow{2}{*}{Method} & \multirow{2}{*}{Backbone} &
        \multicolumn{2}{c}{Unsupervised} & \multicolumn{2}{c}{Linear} \\
        \multicolumn{2}{l}{} & Acc. & mIoU & Acc. & mIoU \\
        \hlineB{2.5}
        MDC~\cite{deepcluster} & R18+FPN & 40.7 & 7.1 & - & - \\
        IIC~\cite{iic} & R18+FPN & 47.9 & 6.4 & - & - \\
        PiCIE~\cite{picie} & R18+FPN & 65.5 & 12.3 & - & - \\
        \hline
        DINO~\cite{dino} & ViT-S/8 & 34.5 & 10.9 & 84.6 & 22.8 \\
        + TransFGU~\cite{transfgu} & ViT-S/8 & 77.9 & 16.8 & - & - \\
        + HP~(Ours) & ViT-S/8 & \textbf{80.1} & \textbf{18.4} & \textbf{91.2} & \textbf{30.6} \\ 
        \hline
        DINO~\cite{dino} & ViT-B/8 & 43.6 & 11.8 & 84.2 & 23.0 \\
        + STEGO~\cite{stego} & ViT-B/8 & 73.2 & \textbf{21.0} & 90.3 & 26.8 \\
        + HP~(Ours) & ViT-B/8 & \textbf{79.5} & 18.4 & \textbf{90.9} & \textbf{33.0} \\ 
        \hlineB{2.5}
    \end{tabular}
    \caption{Experimental results on Cityscapes dataset.}
    \label{Tab.cityscapes}
\end{table}
\endgroup

\begingroup
\setlength{\tabcolsep}{6pt} 
\renewcommand{\arraystretch}{1.0} 
\begin{table}[!t]
    \centering
    \small
    \begin{tabular}{l c c}
        \hlineB{2.5}
        Method & Backbone & Unsup. Acc. \\
        \hlineB{2.5}
        Random CNN~\cite{iic}& VGG11 & 38.2 \\
        K-Means~\cite{sklearn}& VGG11 & 45.7 \\
        SIFT~\cite{sift} & VGG11 &38.2 \\
        ContextPrediction~\cite{context} & VGG11 & 49.6 \\
        CC~\cite{cc} & VGG11 & 63.9 \\
        DeepCluster~\cite{deepcluster} & VGG11 & 41.7 \\
        IIC~\cite{iic} & VGG11 & 65.1 \\
        \hline
        DINO~\cite{dino} & ViT-B/8 & 53.0 \\
        + STEGO~\cite{stego}& ViT-B/8 & 77.0 \\
        + HP~(Ours) & ViT-B/8 & \textbf{82.4}\\
        \hlineB{2.5}
    \end{tabular}
    \caption{Experimental results on Potsdam-3 dataset.}
    \label{Tab.potsdam}
\end{table}
\endgroup

\begin{figure*}[t]
    \centering
    \includegraphics[width=1.0\textwidth]{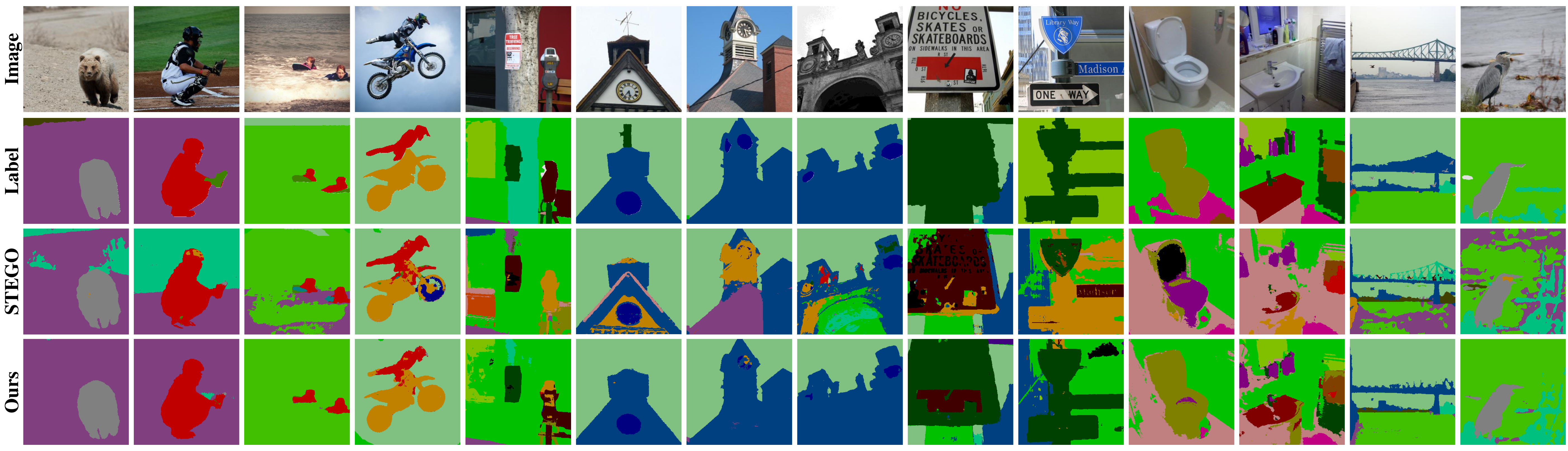}
    \caption{
    Qualitative comparison results of Ours and STEGO on the COCO-stuff dataset with DINO pretrained ViT-S/8 backbone.
    }
    \label{fig_qualitative_result}
\end{figure*}

\subsection{Experimental Results}
We compare our proposed method against the prior techniques for the unsupervised segmentation~\cite{deepcluster, iic, picie, stego, transfgu}.
Most of the results in the result tables are brought from the literature~\cite{picie, stego}.
In Tab.~\ref{Tab.coco}, it is observed that self-supervised models, i.e., DINO and SelfPatch, are already good segmentation predictors with the linear probe, which makes them a new baseline over the prior works for unsupervised segmentation.
Furthermore, we utilize two pretrained backbones with two kinds of architectures to compare with STEGO, in detail.
As reported, our proposed model provides consistent performance improvements over the previous SOTA model in almost all cases on the COCO-stuff dataset.



Results on Cityscapes also show a similar tendency.
As shown in Tab.~\ref{Tab.cityscapes}, ours outperform previous methods except for the mIoU when clustering is used for the evaluation.
For instance, we achieve 8.6\% and 23\% improvements in cluster accuracy and linear mIoU over STEGO, respectively.
For the slight decrease in cluster mIoU with ViT-B/8 architecture, we argue that it is insignificant since the linear probe better describes the quality of representations.
Specifically, clustering evaluation highly depends on the purpose of the dataset so that it is sensitive to the degree of class-specificity as different body parts can be either classified as human or as independent body parts.
However, whereas it is more appropriate to detect each body part independently for unsupervised learning as ours do in Fig.~\ref{fig_selected_positives}, the annotations for general datasets, e.g., COCO-stuff and cityscape, treat these body parts as a human class.
Such circumstances make the clustering evaluation vulnerable to the degree of the class hierarchy.
In contrast, the linear probe projects these features to close proximity if the given label space considers them as a human body.
Thus, we believe the linear probe is a more appropriate measure of representation quality.

Also, we compare the cluster accuracy in Tab.~\ref{Tab.potsdam} on Potsdam-3 dataset.
We have achieved a 7\% boost over STEGO, confirming that ours also performs well even in a completely new domain.
Likewise, our superior results verify the effectiveness of our process of discovering global- and local-hidden positive patches.

\paragraph{Qualitative Results.}
In addition to the quantitative results, we report qualitative results in Fig.~\ref{fig_qualitative_result}.
In comparison to STEGO, we observe that our results include fewer mispredicted pixels throughout the images, while the strength of preserving semantic locality is also validated.
For instance, whereas we find that the predicted label of the wheel is only partially correct in the 4th column, our results are consistent along the neighboring pixels.
These results also demonstrate the superiority of our method.


\section{Ablation Study and Further Analysis}
In this section, we provide ablation studies and an analysis of our model.
Particularly, we explore the contributions of main components and test with varying hyperparameters.
Most experiments for ablation studies are conducted on the COCO-stuff dataset using the DINO pretrained ViT-S/8 model, except for Sec.~\ref{sec.lossprop}, which utilized the ViT-S/16.

\subsection{Importance of the Main Components.}
Tab.~\ref{Tab.ablation} reports the performances when an individual component or various combinations of them are not utilized. 
We found that task-specific GHP and LHP are essential in improving the performance of the unsupervised segmentation task.
Compared to (d) where both the task-specific GHP and LHP are not used, the use of them each leads to 6.5\%~(b) and 11.6\%~(c) improvements.
Also when used together, they bring 16\% of performance boosts~(a).
Furthermore, the importance of preserving symmetricity in selecting GHP can be found by comparing (a) to (e) as its usage boosts 5.9\%,
and consistency regularizer enhances performance by 2.5\% by comparing (a) to (f).
Lastly, (g) shows the performance of naively implemented contrastive learning~(Eq.~\ref{eq_contrastive_loss_self}) with photometric perturbations used in PiCIE~\cite{picie}.
This also confirms the strengths of our proposed method as (a) enhances (g) by 51.3\%.
\begingroup
\setlength{\tabcolsep}{6pt} 
\renewcommand{\arraystretch}{1.0} 
\begin{table}[!t]
    \centering
    \small
    \begin{tabular}{c|c | c |c |c| c|c c}
        \hlineB{2.5}
        \multirow{2}{*}{} & \multicolumn{2}{c|}{GHP} & \multirow{2}{*}{LHP} & \multirow{2}{*}{SA} & \multirow{2}{*}{Reg} & \multicolumn{2}{c}{Unsupervised} \\
        \cline{2-3}
        &TA & TS & & & & Acc. & mIoU \\
        \hlineB{2.5}
        (a)& \checkmark & \checkmark & \checkmark & \checkmark &  \checkmark &  57.2 & 24.6 \\
        (b)&\checkmark & \checkmark & & \checkmark & \checkmark & 52.5 & 23.1 \\ 
        (c)&\checkmark & & \checkmark & \checkmark & \checkmark &  55.0 & 19.1 \\ 
        (d)&\checkmark & & & \checkmark & \checkmark & 49.3 & 20.1 \\ 
        (e)&\checkmark & \checkmark & \checkmark & & \checkmark & 54.0 & 23.6 \\ 
        (f) &\checkmark & \checkmark & \checkmark & \checkmark & & 55.8 & 24.5 \\
        \hline
        (g)&& & & & & 37.8 & 10.4 \\ 
        \hlineB{2.5}
    \end{tabular}
    \caption{Ablation study for each component. GHP, LHP, TA, TS, SA, and Reg denote Global Hidden Positive, Local Hidden Positive, task-agnostic, task-specific, symmetrical assignments, and consistency regularizer, respectively.
    }
    \label{Tab.ablation}
\end{table}
\endgroup

\subsection{Alternatives to Gradient Propagation Strategy}
\label{sec.lossprop}
There can be alternative ways to meet our goal of reflecting semantic consistency between adjacent patches.
As an alternative method of gradient propagation, we simply apply the identical loss to the surrounding patches proportionally to their attention score (i.e., loss propagation).
The results in Tab.~\ref{table_lossprop} show that the loss propagation strategy performs comparably to the gradient propagation strategy. Nonetheless, we observed that the loss propagation approach incurs higher computational costs~($1.2\times$ memory and $3\times$ time). Likewise, our method is implemented considering both effectiveness and efficiency.

\begin{table}[t]
\renewcommand{\arraystretch}{1.0} 
	\centering
	{\footnotesize
        \begin{tabular}{c c c c c}
        \hlineB{2.5}
        \multirow{2}{*}{Method} & \multicolumn{2}{c}{Unsupervised} & \multicolumn{2}{c}{Linear} \\
        \multirow{2}{*}{} & Acc. & mIoU & Acc. & mIoU \\
        \hline
        DINO~+~GradProp & 54.5 & 24.3 & 74.1 & 39.1 \\
        DINO~+~LossProp & 54.7 & 23.2 & 74.3 & 40.5\\
        \hline
        SelfPatch~+~GradProp & 56.1 & 23.2 & 74.9 & 41.3 \\
        SelfPatch~+~LossProp & 54.5 & 22.2 & 75.1 & 41.4 \\
        \hlineB{2.5}
        \end{tabular}
    }
    	\caption{Comparison results between gradient propagation and loss propagation strategies.}
	\label{table_lossprop}
\end{table}

\begin{figure}[t]
    \centering
    \includegraphics[width=0.47\textwidth]{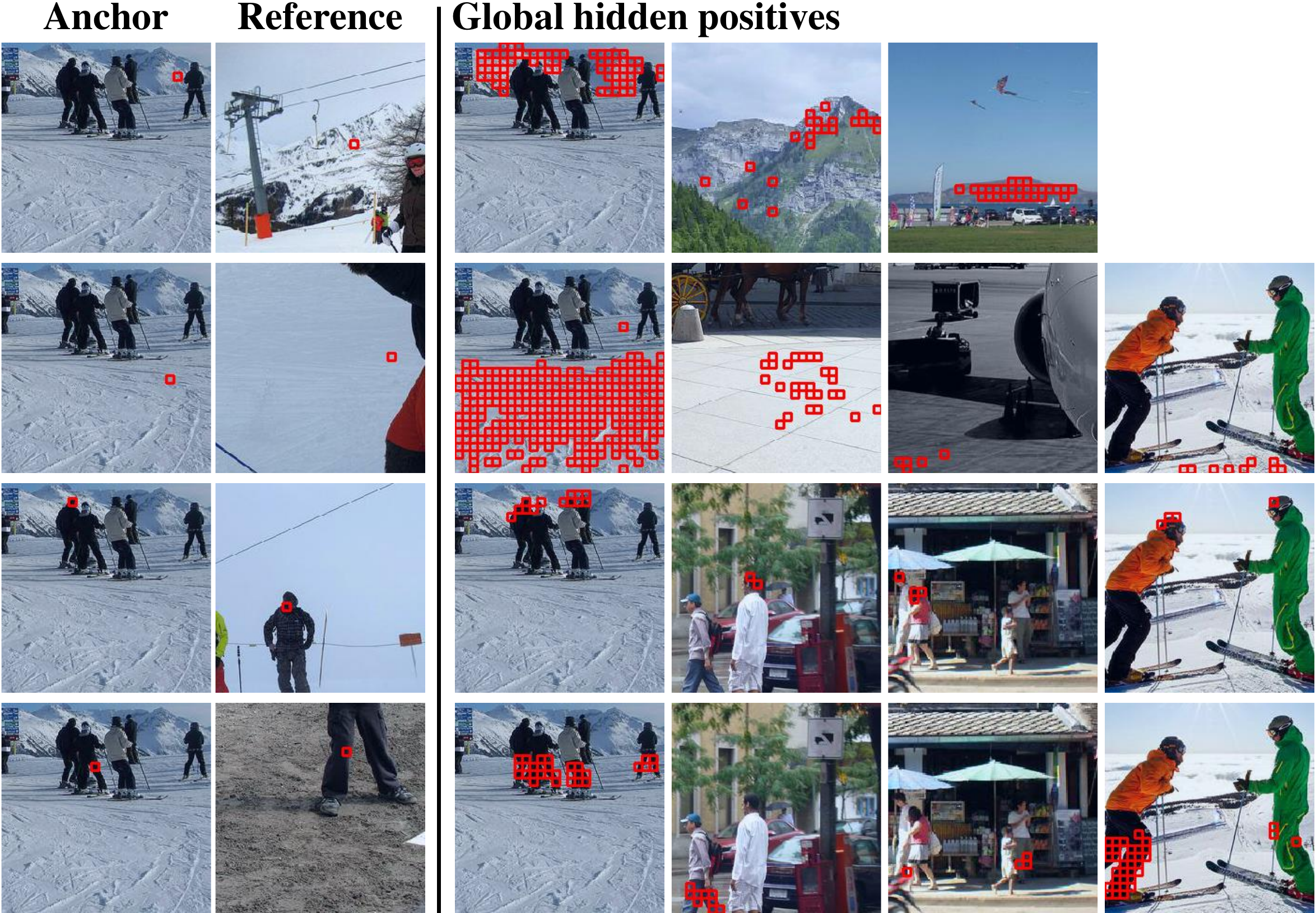}
    \caption{
    Discovered patches by our GHP selection process. From the left to right columns, red boxes indicate the anchors, the closest patches within the task-agnostic reference pool~(reference), and GHP sets chosen in the mini-batch, respectively.
    }
    \label{fig_selected_positives}
\end{figure}

\subsection{Visualization}
\paragraph{Discovered GHP.}
To demonstrate the effectiveness of our GHP selection process, we visualize the selected GHP sets for different anchors in a single image in Fig.~\ref{fig_selected_positives}.
First, we observe that the corresponding reference point in the second column is semantically correlated with the anchor patch.
Also, the obtained GHP sets verify the appropriateness of the use of reference points as each anchor's criterion as they are capable of precisely discovering the semantically-similar positives.
For example, we find that all the anchors, reference points, and GHP sets have the same semantic labels in the first and the second rows~(solid~(mountain) and ground~(snow)).
More intriguing results are in the third and fourth rows where we find that the GHP selection process distinguishes the body parts in a much more fine-grained manner than the given annotation, i.e., person.
These results imply that the designed GHP selection is well-designed and capable of capturing detailed semantic contexts.



\subsection{Robustness to Hyperparameters}

\begin{figure}[t]
    \centering
    \includegraphics[width=0.47\textwidth]{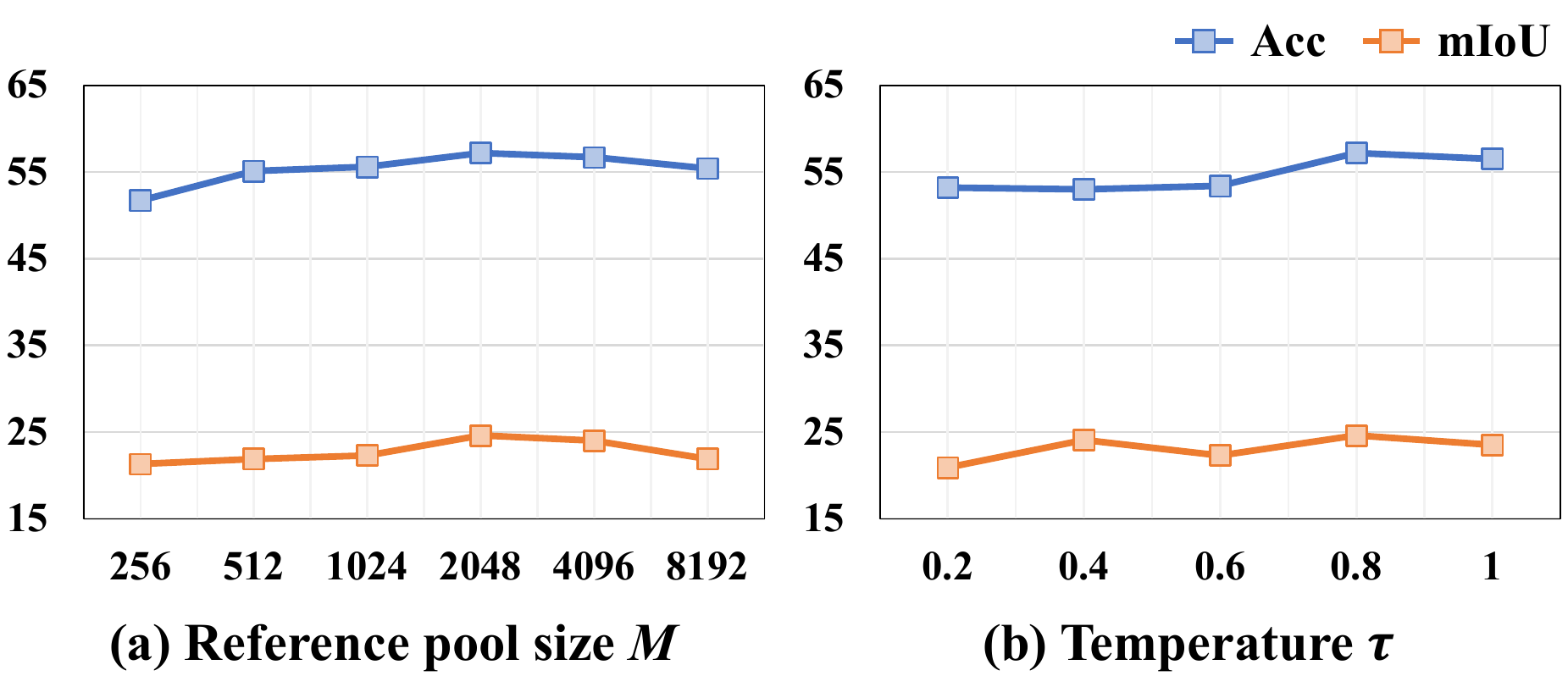}
    \caption{
    Ablation studies on hyperparameters.
    }
    \label{fig_ablation_hyperparameters}
\end{figure}

In this subsection, we enumerate our use of hyperparameters in Tab.~\ref{Tab.hyperparamters} and conduct an ablation study.
Overall, our proposed training scheme is robust to hyperparameters as Fig.~\ref{fig_ablation_hyperparameters} supports the claim.
Below, we illustrate the influences of each parameter.
Despite the slight drop in performance from time to time, it is insignificant since the change in performance is very marginal and results are consistent.

\noindent
\paragraph{The Number of Data in the Reference Pool.}
Fig.~\ref{fig_ablation_hyperparameters}~(a) shows the performances with varying numbers of data $M$ in the reference pool.
Generally, the reference pool is not very vulnerable to its size, unless the capacity is either too small or too large.
When the reference pool is skinny, it may not be sufficient to represent all kinds of semantics present in the dataset.
In other words, the small reference pool may induce a biased criterion for each anchor which may incur biased training.
On the other hand, when the reference pool is too large, a tight threshold~($c_i$) could be derived, which can interfere with gathering GHP.



\begingroup
\setlength{\tabcolsep}{6pt} 
\renewcommand{\arraystretch}{1.0} 
\begin{table}[t!]
    \centering
    \small
    \begin{tabular}{l c c}
        \hlineB{2.5}
        Dataset & $M$ & $\tau$ \\
        \hlineB{2.5}
        COCO-stuff & 2048 & 0.8 \\
        Cityscapes & 2048 & 0.6 \\
        Potsdam-3 & 1024 & 0.4 \\
        \hlineB{2.5}
    \end{tabular}
    \caption{Hyperparameter used for each dataset.}
    \label{Tab.hyperparamters}
\end{table}
\endgroup

\noindent
\paragraph{Temperature Parameter.}
The temperature parameter $\tau$ is the scaling parameter to manipulate the sensitivity of the contrastive loss. 
If $\tau$ gets bigger, the objective function becomes robust to the difference between the similarity of the positive and negative samples.
On the other hand, when $\tau$ gets lower, the embedding distribution is likely to be more uniform~\cite{wang2021understanding}.
Although the results do not fluctuate much, we observe that training uniformly distributed embedding space leads to a slight performance drop in Fig.~\ref{fig_ablation_hyperparameters}~(b).

\section{Conclusion}
In this paper, we introduced a novel unsupervised semantic segmentation method by discovering and leveraging two types of hidden positives, global hidden positive~(GHP) and local hidden positive~(LHP), to learn rich semantic information with local consistency. 
First, anchor-dependent GHP comprises task-agnostic and task-specific positive sets which are used to tailor the contrastive learning for the unsupervised semantic segmentation task.
Whereas the task-agnostic features are collected to guide the initial training, task-specific features are progressively engaged to learn the task-specific semantics information.
Moreover, under the inherent premise that the adjacent patches are likely to be semantically similar, we propagate the loss gradient to the surrounding patches in proportion to their attention scores.
This encourages the semantically similar peripheral patches to have the same objective as the anchor, resulting in semantic consistency between adjacent patches unless they belong to different objects.
Finally, our proposed method achieves new state-of-the-art results in various datasets.

\vspace{5pt}
\noindent\textbf{Acknowledgements.} This work was supported in part by MSIT/IITP (No. 2022-0-00680, 2019-0-00421, 2020-0-01821, 2021-0-02068), and MSIT\&KNPA/KIPoT (Police Lab 2.0, No. 210121M06).

{\small
\bibliographystyle{ieee_fullname}
\bibliography{egbib}

\begin{thebibliography}{10}\itemsep=-1pt

\bibitem{coco}
Holger Caesar, Jasper Uijlings, and Vittorio Ferrari.
\newblock Coco-stuff: Thing and stuff classes in context.
\newblock In {\em Proceedings of the IEEE conference on computer vision and
  pattern recognition}, pages 1209--1218, 2018.

\bibitem{deepcluster}
Mathilde Caron, Piotr Bojanowski, Armand Joulin, and Matthijs Douze.
\newblock Deep clustering for unsupervised learning of visual features.
\newblock In {\em Proceedings of the European conference on computer vision
  (ECCV)}, pages 132--149, 2018.

\bibitem{swav}
Mathilde Caron, Ishan Misra, Julien Mairal, Priya Goyal, Piotr Bojanowski, and
  Armand Joulin.
\newblock Unsupervised learning of visual features by contrasting cluster
  assignments.
\newblock {\em Advances in Neural Information Processing Systems},
  33:9912--9924, 2020.

\bibitem{dino}
Mathilde Caron, Hugo Touvron, Ishan Misra, Herv{\'e} J{\'e}gou, Julien Mairal,
  Piotr Bojanowski, and Armand Joulin.
\newblock Emerging properties in self-supervised vision transformers.
\newblock In {\em Proceedings of the IEEE/CVF International Conference on
  Computer Vision}, pages 9650--9660, 2021.

\bibitem{deeplab}
Liang-Chieh Chen, George Papandreou, Iasonas Kokkinos, Kevin Murphy, and Alan~L
  Yuille.
\newblock Semantic image segmentation with deep convolutional nets and fully
  connected crfs.
\newblock {\em ICLR}, 2015.

\bibitem{simclr}
Ting Chen, Simon Kornblith, Mohammad Norouzi, and Geoffrey Hinton.
\newblock A simple framework for contrastive learning of visual
  representations.
\newblock In {\em International conference on machine learning}, pages
  1597--1607. PMLR, 2020.

\bibitem{simsiam}
Xinlei Chen and Kaiming He.
\newblock Exploring simple siamese representation learning.
\newblock In {\em Proceedings of the IEEE/CVF Conference on Computer Vision and
  Pattern Recognition}, pages 15750--15758, 2021.

\bibitem{picie}
Jang~Hyun Cho, Utkarsh Mall, Kavita Bala, and Bharath Hariharan.
\newblock Picie: Unsupervised semantic segmentation using invariance and
  equivariance in clustering.
\newblock In {\em Proceedings of the IEEE/CVF Conference on Computer Vision and
  Pattern Recognition}, pages 16794--16804, 2021.

\bibitem{cityscapes}
Marius Cordts, Mohamed Omran, Sebastian Ramos, Timo Rehfeld, Markus Enzweiler,
  Rodrigo Benenson, Uwe Franke, Stefan Roth, and Bernt Schiele.
\newblock The cityscapes dataset for semantic urban scene understanding.
\newblock In {\em Proceedings of the IEEE conference on computer vision and
  pattern recognition}, pages 3213--3223, 2016.

\bibitem{context}
Carl Doersch, Abhinav Gupta, and Alexei~A Efros.
\newblock Unsupervised visual representation learning by context prediction.
\newblock In {\em Proceedings of the IEEE international conference on computer
  vision}, pages 1422--1430, 2015.

\bibitem{danet}
Jun Fu, Jing Liu, Haijie Tian, Yong Li, Yongjun Bao, Zhiwei Fang, and Hanqing
  Lu.
\newblock Dual attention network for scene segmentation.
\newblock In {\em Proceedings of the IEEE/CVF conference on computer vision and
  pattern recognition}, pages 3146--3154, 2019.

\bibitem{rotnet}
Spyros Gidaris, Praveer Singh, and Nikos Komodakis.
\newblock Unsupervised representation learning by predicting image rotations.
\newblock In {\em International Conference on Learning Representations}, 2018.

\bibitem{byol}
Jean-Bastien Grill, Florian Strub, Florent Altch{\'e}, Corentin Tallec, Pierre
  Richemond, Elena Buchatskaya, Carl Doersch, Bernardo Avila~Pires, Zhaohan
  Guo, Mohammad Gheshlaghi~Azar, et~al.
\newblock Bootstrap your own latent-a new approach to self-supervised learning.
\newblock {\em Advances in neural information processing systems},
  33:21271--21284, 2020.

\bibitem{stego}
Mark Hamilton, Zhoutong Zhang, Bharath Hariharan, Noah Snavely, and William~T.
  Freeman.
\newblock Unsupervised semantic segmentation by distilling feature
  correspondences.
\newblock In {\em International Conference on Learning Representations}, 2022.

\bibitem{moco}
Kaiming He, Haoqi Fan, Yuxin Wu, Saining Xie, and Ross Girshick.
\newblock Momentum contrast for unsupervised visual representation learning.
\newblock In {\em Proceedings of the IEEE/CVF conference on computer vision and
  pattern recognition}, pages 9729--9738, 2020.

\bibitem{hendrycks2019using}
Dan Hendrycks, Mantas Mazeika, Saurav Kadavath, and Dawn Song.
\newblock Using self-supervised learning can improve model robustness and
  uncertainty.
\newblock {\em Advances in neural information processing systems}, 32, 2019.

\bibitem{segsort}
Jyh-Jing Hwang, Stella~X Yu, Jianbo Shi, Maxwell~D Collins, Tien-Ju Yang, Xiao
  Zhang, and Liang-Chieh Chen.
\newblock Segsort: Segmentation by discriminative sorting of segments.
\newblock In {\em Proceedings of the IEEE/CVF International Conference on
  Computer Vision}, pages 7334--7344, 2019.

\bibitem{cc}
Phillip Isola, Daniel Zoran, Dilip Krishnan, and Edward~H Adelson.
\newblock Learning visual groups from co-occurrences in space and time.
\newblock {\em arXiv preprint arXiv:1511.06811}, 2015.

\bibitem{iic}
Xu Ji, Joao~F Henriques, and Andrea Vedaldi.
\newblock Invariant information clustering for unsupervised image
  classification and segmentation.
\newblock In {\em Proceedings of the IEEE/CVF International Conference on
  Computer Vision}, pages 9865--9874, 2019.

\bibitem{supcon}
Prannay Khosla, Piotr Teterwak, Chen Wang, Aaron Sarna, Yonglong Tian, Phillip
  Isola, Aaron Maschinot, Ce Liu, and Dilip Krishnan.
\newblock Supervised contrastive learning.
\newblock {\em Advances in Neural Information Processing Systems},
  33:18661--18673, 2020.

\bibitem{crf}
Philipp Kr{\"a}henb{\"u}hl and Vladlen Koltun.
\newblock Efficient inference in fully connected crfs with gaussian edge
  potentials.
\newblock {\em Advances in neural information processing systems}, 24, 2011.

\bibitem{ema}
Samuli Laine and Timo Aila.
\newblock Temporal ensembling for semi-supervised learning.
\newblock {\em ICLR}, 2017.

\bibitem{ficklenet}
Jungbeom Lee, Eunji Kim, Sungmin Lee, Jangho Lee, and Sungroh Yoon.
\newblock Ficklenet: Weakly and semi-supervised semantic image segmentation
  using stochastic inference.
\newblock In {\em Proceedings of the IEEE/CVF Conference on Computer Vision and
  Pattern Recognition}, pages 5267--5276, 2019.

\bibitem{sift}
David~G Lowe.
\newblock Object recognition from local scale-invariant features.
\newblock In {\em Proceedings of the seventh IEEE international conference on
  computer vision}, volume~2, pages 1150--1157. Ieee, 1999.

\bibitem{pirl}
Ishan Misra and Laurens van~der Maaten.
\newblock Self-supervised learning of pretext-invariant representations.
\newblock In {\em Proceedings of the IEEE/CVF Conference on Computer Vision and
  Pattern Recognition}, pages 6707--6717, 2020.

\bibitem{lorot}
WonJun Moon, Ji-Hwan Kim, and Jae-Pil Heo.
\newblock Tailoring self-supervision for supervised learning.
\newblock In {\em European Conference on Computer Vision}, pages 346--364.
  Springer, 2022.

\bibitem{sklearn}
Fabian Pedregosa, Ga{\"e}l Varoquaux, Alexandre Gramfort, Vincent Michel,
  Bertrand Thirion, Olivier Grisel, Mathieu Blondel, Peter Prettenhofer, Ron
  Weiss, Vincent Dubourg, et~al.
\newblock Scikit-learn: Machine learning in python.
\newblock {\em the Journal of machine Learning research}, 12:2825--2830, 2011.

\bibitem{unet}
Olaf Ronneberger, Philipp Fischer, and Thomas Brox.
\newblock U-net: Convolutional networks for biomedical image segmentation.
\newblock In {\em International Conference on Medical image computing and
  computer-assisted intervention}, pages 234--241. Springer, 2015.

\bibitem{segmenter}
Robin Strudel, Ricardo Garcia, Ivan Laptev, and Cordelia Schmid.
\newblock Segmenter: Transformer for semantic segmentation.
\newblock In {\em Proceedings of the IEEE/CVF International Conference on
  Computer Vision}, pages 7262--7272, 2021.

\bibitem{maskcontrast}
Wouter Van~Gansbeke, Simon Vandenhende, Stamatios Georgoulis, and Luc Van~Gool.
\newblock Unsupervised semantic segmentation by contrasting object mask
  proposals.
\newblock In {\em Proceedings of the IEEE/CVF International Conference on
  Computer Vision}, pages 10052--10062, 2021.

\bibitem{wang2021understanding}
Feng Wang and Huaping Liu.
\newblock Understanding the behaviour of contrastive loss.
\newblock In {\em Proceedings of the IEEE/CVF conference on computer vision and
  pattern recognition}, pages 2495--2504, 2021.

\bibitem{eisnet}
Shujun Wang, Lequan Yu, Caizi Li, Chi-Wing Fu, and Pheng-Ann Heng.
\newblock Learning from extrinsic and intrinsic supervisions for domain
  generalization.
\newblock In {\em European Conference on Computer Vision}, pages 159--176.
  Springer, 2020.

\bibitem{seam}
Yude Wang, Jie Zhang, Meina Kan, Shiguang Shan, and Xilin Chen.
\newblock Self-supervised equivariant attention mechanism for weakly supervised
  semantic segmentation.
\newblock In {\em Proceedings of the IEEE/CVF Conference on Computer Vision and
  Pattern Recognition}, pages 12275--12284, 2020.

\bibitem{segformer}
Enze Xie, Wenhai Wang, Zhiding Yu, Anima Anandkumar, Jose~M Alvarez, and Ping
  Luo.
\newblock Segformer: Simple and efficient design for semantic segmentation with
  transformers.
\newblock {\em Advances in Neural Information Processing Systems},
  34:12077--12090, 2021.

\bibitem{mctformer}
Lian Xu, Wanli Ouyang, Mohammed Bennamoun, Farid Boussaid, and Dan Xu.
\newblock Multi-class token transformer for weakly supervised semantic
  segmentation.
\newblock In {\em Proceedings of the IEEE/CVF Conference on Computer Vision and
  Pattern Recognition}, pages 4310--4319, 2022.

\bibitem{nsrom}
Yazhou Yao, Tao Chen, Guo-Sen Xie, Chuanyi Zhang, Fumin Shen, Qi Wu, Zhenmin
  Tang, and Jian Zhang.
\newblock Non-salient region object mining for weakly supervised semantic
  segmentation.
\newblock In {\em Proceedings of the IEEE/CVF Conference on Computer Vision and
  Pattern Recognition}, pages 2623--2632, 2021.

\bibitem{transfgu}
Zhaoyuan Yin, Pichao Wang, Fan Wang, Xianzhe Xu, Hanling Zhang, Hao Li, and
  Rong Jin.
\newblock Transfgu: a top-down approach to fine-grained unsupervised semantic
  segmentation.
\newblock In {\em European Conference on Computer Vision}, pages 73--89.
  Springer, 2022.

\bibitem{ocr}
Yuhui Yuan, Xilin Chen, and Jingdong Wang.
\newblock Object-contextual representations for semantic segmentation.
\newblock In {\em European conference on computer vision}, pages 173--190.
  Springer, 2020.

\bibitem{selfpatch}
Sukmin Yun, Hankook Lee, Jaehyung Kim, and Jinwoo Shin.
\newblock Patch-level representation learning for self-supervised vision
  transformers.
\newblock In {\em Proceedings of the IEEE/CVF Conference on Computer Vision and
  Pattern Recognition}, pages 8354--8363, 2022.

\bibitem{supseg}
Xiangyun Zhao, Raviteja Vemulapalli, Philip~Andrew Mansfield, Boqing Gong,
  Bradley Green, Lior Shapira, and Ying Wu.
\newblock Contrastive learning for label efficient semantic segmentation.
\newblock In {\em Proceedings of the IEEE/CVF International Conference on
  Computer Vision}, pages 10623--10633, 2021.

\bibitem{setr}
Sixiao Zheng, Jiachen Lu, Hengshuang Zhao, Xiatian Zhu, Zekun Luo, Yabiao Wang,
  Yanwei Fu, Jianfeng Feng, Tao Xiang, Philip~HS Torr, et~al.
\newblock Rethinking semantic segmentation from a sequence-to-sequence
  perspective with transformers.
\newblock In {\em Proceedings of the IEEE/CVF conference on computer vision and
  pattern recognition}, pages 6881--6890, 2021.

\bibitem{sessd}
Wu Zheng, Weiliang Tang, Li Jiang, and Chi-Wing Fu.
\newblock Se-ssd: Self-ensembling single-stage object detector from point
  cloud.
\newblock In {\em Proceedings of the IEEE/CVF Conference on Computer Vision and
  Pattern Recognition}, pages 14494--14503, 2021.

\end{thebibliography}
}

\end{document}